\documentclass[conference]{IEEEtran}
\IEEEoverridecommandlockouts
\usepackage{cite}
\usepackage{amsmath,amssymb,amsfonts}
\usepackage{algorithmic}
\usepackage{graphicx}
\usepackage{textcomp}
\usepackage{xcolor}
\usepackage{cite}
\usepackage{hyperref}
\usepackage{booktabs}
\usepackage{float}
\usepackage[nottoc, numbib]{tocbibind}
\usepackage{caption}
\usepackage{booktabs}

\def\BibTeX{{\rm B\kern-.05em{\sc i\kern-.025em b}\kern-.08em
    T\kern-.1667em\lower.7ex\hbox{E}\kern-.125emX}}
\begin{document}

\title{LiqD: A Dynamic Liquid Level Detection Model under Tricky Small Containers}

\author{\IEEEauthorblockN{1\textsuperscript{st} Yukun Ma}
\IEEEauthorblockA{\textit{School of Electronic and Information Engineering} \\
\textit{Beijing Jiatong University}\\
Beijing, China \\
21721018@bjtu.edu.cn}
\and
\IEEEauthorblockN{2\textsuperscript{nd} Zikun Mao}
\IEEEauthorblockA{\textit{School of Electronic and Information Engineering} \\
\textit{Beijing Jiatong University}\\
Beijing, China \\
21721063@bjtu.edu.cn}
\and
}

\maketitle

\begin{abstract}
In daily life and industrial production, it is crucial to accurately detect changes in liquid level in containers. Traditional contact measurement methods have some limitations, while emerging non-contact image processing technology shows good application prospects. This paper proposes a container dynamic liquid level detection model based on U²-Net. This model uses the SAM model to generate an initial data set, and then evaluates and filters out high-quality pseudo-label images through the SemiReward framework to build an exclusive data set. The model uses U²-Net to extract mask images of containers from the data set, and uses morphological processing to compensate for mask defects. Subsequently, the model calculates the grayscale difference between adjacent video frame images at the same position, segments the liquid level change area by setting a difference threshold, and finally uses a lightweight neural network to classify the liquid level state. This approach not only mitigates the impact of intricate surroundings, but also reduces the demand for training data, showing strong robustness and versatility. A large number of experimental results show that the proposed model can effectively detect the dynamic liquid level changes of the liquid in the container, providing a novel and efficient solution for related fields.
\end{abstract}


\section{Introduction}
liquid level detection technology in containers plays a vital role in daily life. It not only prevents liquid overflow in home kitchens and ensures cooking safety, but also monitors the amount of liquid in storage tanks and reactors in the industrial field to ensure production processes. Smooth and safe. Also in construction, liquid level detection is used to monitor liquid levels in tunnels and underground facilities to prevent flooding and structural damage. Scenarios like this are widely used.

To accurately monitor liquid levels, traditional contact measurement methods like float gauges and pressure transmitters\cite{singh2018review} offer high measurement accuracy but have certain limitations. These methods require the measuring element to be directly immersed in the liquid, making them unsuitable for harsh environments involving highly corrosive substances, extreme temperatures, or high pressures.

In recent years, some non-contact remote measurement technologies have rapidly advanced, such as liquid level measurement systems based on radar and sonar principles\cite{mohindru2023development}. These novel techniques eliminate the necessity for physical contact with the liquid being measured, offer a wide measurement range, and highly adapt to different environments. However, they also face challenges like relatively high system costs and strict requirements on environmental conditions (e.g., temperature, pressure).

With continuous advancements in computer vision and image processing, image-based liquid level detection methods are increasingly emerging and attracting widespread industry attention. Traditional image algorithms have proposed numerous liquid level detection methods using image shooting and processing to obtain liquid level conditions through spatial mathematical relationships\cite{wang2009liquid}. These methods achieved convincing results over a decade ago. However, the application of deep learning in image processing has ushered in a new era. For liquid level detection in large scenes like lakes and reservoirs, substantial advancements have been achieved\cite{YYKX202203007}\cite{JSJZ202103088}\cite{BWTS202304001009}\cite{lin2018automatic}\cite{DNXJ202203006}. For example, Fang et al.\cite{BWTS202304001009} used YOLOv4 to accurately locate liquid gauge scale characters, then DeepLabv3+ to precisely segment the junction area between the gauge and liquid body, and finally extracted liquid levels and calculated actual values using image processing techniques. Sun et al.\cite{YYKX202203007} achieved high-precision, real-time liquid level monitoring through steps like image preprocessing, edge detection, affine transformation correction, keyword positioning, and edge projection. Xia et al.\cite{JSJZ202103088} improved the superpixel and graph cutting algorithm, then performed liquid level detection based on the semantic segmentation network technology of U-net. Zhang et al.\cite{DNXJ202203006} proposed a liquid level height difference prediction method based on digital image processing by using a digital camera to capture a top view of the container, then performing image preprocessing, edge detection, and ellipse fitting to calculate the liquid level and distance from the container top.

These methods have improved accuracy, generalization ability, and environmental adaptability but still face challenges and bottlenecks. Firstly, existing research mainly focuses on large liquid bodies, lacking relevant technology accumulation for small container scenarios. Secondly, most algorithms have high training data requirements, resulting in poor generalization capabilities when applied to different environments. Furthermore, complex environments introduce interference like lighting and occlusion, affecting detection accuracy. Mitigating the influence of environmental factors remains a critical challenge. Finally, for dynamically changing liquid levels, accurate and stable detection is challenging due to factors like fluctuations, and existing methods lack modeling and analysis of dynamic processes. All these challenges await further breakthroughs and research.

Based on the above analysis, we proposed a new visual processing method for dynamic liquid level changes in containers, greatly addressing issues of high sample requirements, complex environmental influences, and limited detection scene sizes. Our main contributions are threefold:

\begin{itemize}
    \item We construct a dedicated dataset using the SAM model and evaluate it through the SemiReward framework to obtain a standardized and specialized dataset.
    
    \item By employing U²-Net for salient object extraction, we obtain the container mask, focusing the analysis solely on the liquid surface within the container image. This not only greatly mitigates interference from external environments but also shifts the detection emphasis toward subtle changes in small-scale features within the image.
    
    \item We adopt image morphological methods to significantly improve the quality of suboptimal masks, resulting in more distinct and smooth boundaries.
\end{itemize}

\section{Related Works}
\subsection{SAM Model}
SAM~\cite{kirillov2023segment} represents an innovative deep learning architecture designed to efficiently segment arbitrary image content through a prompt-based segmentation task. This model can generate precise segmentation masks in real-time, without the need for specific task training, by utilizing flexible prompts such as points, bounding boxes, and text. SAM relies on a large-scale dataset named SA-1B, which includes over 1.1 billion auto-generated masks, ensuring the model's generalization across diverse scenes. The zero-shot transfer learning capabilities of SAM have demonstrated remarkable performance across multiple downstream tasks, marking a significant breakthrough in the field of image segmentation.

It's noteworthy that SAM has learned a universal ability for object recognition and segmentation, thus its exceptional performance is not confined to specific object categories. Whether dealing with a single target or multiple targets of the same or different categories, SAM accurately segments them. This versatility positions SAM for a wide range of applications, such as interactive image editing, general object segmentation, and visual question answering, among others. Beyond segmentation quality, another major advantage of the SAM model is its computational efficiency. With no need for time-consuming task-specific fine-tuning, SAM can respond to user prompts in real-time, rapidly producing segmentation outcomes, thereby facilitating downstream visual tasks and offering an excellent user interaction experience.

SAM's image segmentation capabilities and prompt adaptability guide our container mask creation, creating a foundational dataset for model training. Its scene generalization lets us tackle various container types, broadening our method's scope. While SAM presents real-time interaction, we use it for data creation, not full liquid level detection. To improve dataset reliability, we also integrate SemiReward for mask quality refinement.

\subsection{U²-Net}
The U²-Net\cite{qin2020u2} architecture is a deep learning framework specifically tailored for salient object detection (SOD) tasks. Its core innovation lies in the unique nested U-shaped structure, which effectively captures rich contextual information at different scales. The architecture utilizes Residual U-blocks (RSUs) at each stage to extract multi-scale features while maintaining high-resolution feature maps. The clever design of the RSUs enhances the network's depth without significantly increasing computational costs, allowing U²-Net to be trained from scratch without relying on pre-trained image classification backbones. This design not only improves SOD performance but also computational efficiency, providing a novel and efficient solution for the SOD domain. Unlike traditional methods that depend on pre-trained backbones, U²-Net's ability to train from zero showcases performance comparable to or even better than the current state-of-the-art. And the training loss $L$ from\cite{qin2020u2} is defined as:
\begin{equation}
\label{equ1}
L = \sum_{m=1}^{M} w_{side}^{(m)} l_{side}^{(m)} + w_{fuse} l_{fuse}
\end{equation}

\noindent where $M$ is the number of side-output saliency maps, $w_{side}^{(m)}$ is the weight of the $m$th side-output loss, $l_{side}^{(m)}$ is the loss of the $m$th side-output saliency map, $w_{fuse}$ is the weight of the fusion output loss, and $l_{fuse}$ is the loss of the final fusion output saliency map. Each side-output loss $l_{side}^{(m)}$ is computed using the binary cross-entropy loss from\cite{qin2020u2} as shown below:

\begin{equation}
l=\sum_{(r, c)}^{(H, W)}\left[P_{G(r, c)} \log P_{S(r, c)}+\left(1-P_{G(r, c)}\right) \log \left(1-P_{S(r, c)}\right)\right]
\end{equation}

\noindent where $(r,c)$ are the pixel coordinates, $(H,W)$ is the image size in height and width, $P_G(r,c)$ denotes the pixel values of the ground truth, and $P_S(r,c)$ denotes the pixel values of the predicted saliency probability map. The training process tries to minimize the overall loss $L$ of \eqref{equ1}. In the testing process, the fusion output $l_{fuse}$ used is chosen as the final saliency map.

U²-Net's hierarchical U-shaped architecture and RSUs inform our approach, allowing us to enhance container segmentation precision without increasing computational demands. Its train-from-zero approach enables us to create models tailored for specific container data, deviating from U²-Net's general SOD focus. We've adapted U²-Net for container segmentation by adjusting training data, loss functions, and adding morphological processing to better suit liquid level detection tasks.
\subsection{Bottleneck in Hand-crafted Design}
\subsubsection{Morphological Compensation}
\ 

In the process of image analysis, defective images are commonly encountered. To address this issue, Vizilter et al.\cite{vizilter2015morphological} employed morphological image analysis to solve the problems of change detection and shape matching in images, which is similar to the idea of using morphological operations for image restoration as described by Raid et al.\cite{raid2014image}. By adopting this method, defects can be compensated for by filling holes and connecting broken regions in the image.

Firstly, a structuring element needs to be defined, which specifies the shape and size of the morphological operation. In this study, we chose to use an elliptical structuring element with a size of 5x5 pixels. Morphological closing operation, which consists of dilation followed by erosion, is then applied to the current binary image to fill small holes and connect broken regions[1]. Based on this theory, the following equation from\cite{raid2014image} can be derived:

\begin{equation}
A\oplus B=\left\{ x,y\left| \left( B \right) _{xy}\cap A\ne \oslash \right. \right\} 
\end{equation}
\noindent where $(B)_{xy}$ denotes the translation of the structuring element B such that its origin is at $(x,y)$. The output pixel $(x,y)$ is set to 1 if the intersection of the translated $B$ with the set A is non-empty, otherwise it is set to 0.

Erosion can "shrink" the target region, essentially causing the image boundaries to contract. It can be used to eliminate small, insignificant targets. The equation for erosion from\cite{raid2014image} is expressed as:
\begin{equation}
A\ominus B=\left\{ x,y\left| \left( B \right) _{xy}\subseteq A \right. \right\} 
\end{equation}
\noindent where $(B)_{xy}$ denotes the translation of the structuring element $B$ such that its origin is at $(x,y)$. The output pixel $(x,y)$ is set to 1 if the translated $B$ is completely contained within the set $A$, otherwise it is set to 0. This equation represents the erosion of A by the structuring element $B$.

\subsubsection{Grayscale Value Conversion}
\ 

Most of the images in this study are in color format, but the color information is not highly relevant. Therefore, it is crucial to introduce grayscale conversion to obtain meaningful numerical values. In terms of grayscale conversion methods, Saravanan\cite{saravanan2010color} proposed a novel algorithm that addresses the contrast, sharpness, shadows, and structure of the image. This algorithm approximates, reduces, and adds to the chromaticity and luminosity of the RGB values. The formula from\cite{saravanan2010color} is as follows:
\begin{gather}
Y=0.299R+0.587G+0.114B \notag \\
U=0.565(B-Y) \notag \\
V=0.713(R-Y) \notag \\
I_1=(R/3+G/3+B/3+U+V)/4
\end{gather}
where $Y$ represents luminance, while U and V represent chrominance. The calculation of $Y$ is based on the weighted sum of RGB components, while the calculation of $U$ and $V$ is based on the differences between red, green, blue, and luminance. The intensity value ($I_1$) is computed by taking the average of the RGB components, adding the $U$ and $V$ components, and dividing the sum by 4.

Traditional grayscale image algorithms are not specifically tailored for classification purposes. In the context of image classification, Güneş et al.\cite{gunecs2016optimizing} proposed a novel color-to-grayscale conversion method based on Genetic Algorithm (GA). By utilizing GA, the conversion coefficients for color images are optimized to generate grayscale images with enhanced discriminative features, aiming to reduce errors in image classification problems. The formula from\cite{gunecs2016optimizing} is as follows:
\begin{gather}
r^{\prime}=r/(r+g+b) \notag \\
g^{\prime}=g/(r+g+b) \notag \\
b^{\prime}=b/(r+g+b) \notag \\
I_2=r^{\prime}×R+g^{\prime}×G+b^{\prime}×B
\end{gather}


Integrating the above two methods, the final intensity value $I$ is obtained by adding $I_1$ and $I_2$ through the weighted proportional coefficients $\alpha$ and $\beta$ using the equation:
\begin{equation}
I = \alpha \cdot I_1 + \beta \cdot I_2
\end{equation}
where $\alpha$ and $\beta$ are weighting coefficients satisfying $\alpha + \beta = 1$. $I_1$ takes into account visual factors such as brightness, chromaticity, and contrast, while $I_2$ emphasizes discriminative power for classification. The two methods are complementary to each other. By employing a weighted fusion approach, the visual quality can be enhanced while simultaneously taking classification performance into account.

\section{Method}
\begin{figure*}
    \centering
    \includegraphics[width=1\linewidth]{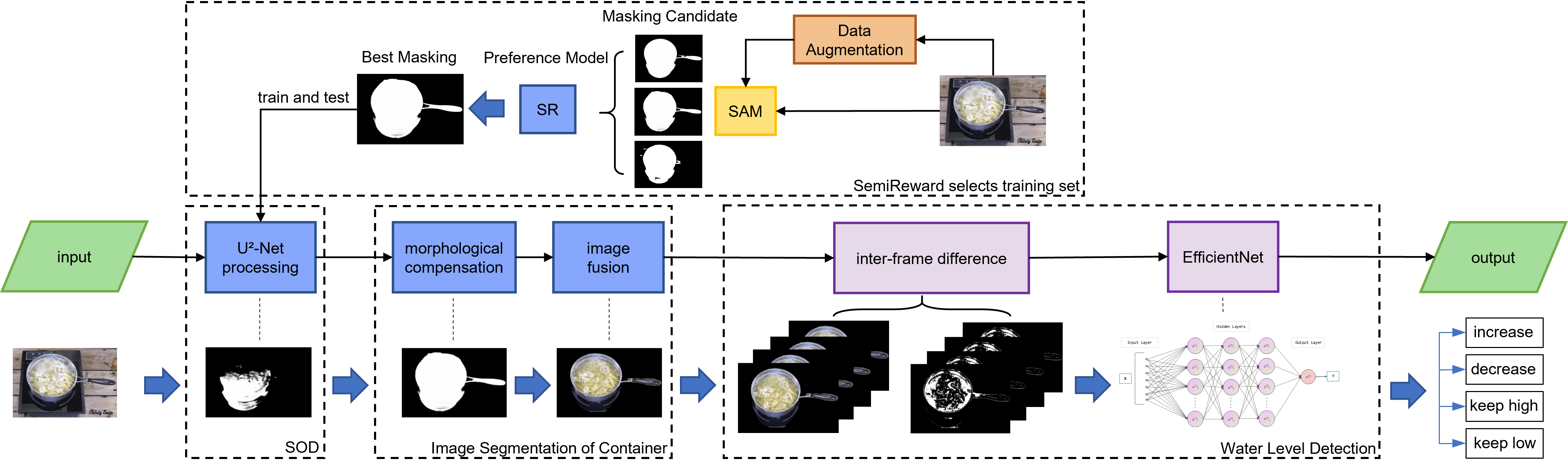}
    \caption{Overall framework}
    \label{fig:overall framework workflow}
\end{figure*}

Based on the algorithm analysis mentioned above, we propose an overall framework workflow as illustrated in Fig.~\ref{fig:overall framework workflow}. The algorithm consists of four core modules: Data engine construction, prominent object extraction from the container, morphological completion of the container shape, and calculation of the height difference in the container for liquid level detection.
\subsection{Construct Data Engine}
For the data engine approach, the core key is how to evaluate and filter labels and how to generate more label candidates of different qualities. SemiReward (SR)~\cite{li2024semireward} has proposed an effective pseudo-label screening method for classification and regression tasks in the past. We modified this method to make it a method that can evaluate our masking. We use common Masking evaluation as The metric that can be learned allows the trained SR model to start evaluating and screening Masking. At the same time, data amplification is performed using methods such as noise addition and the most advanced mix-up~\cite{liu2021automix} to ultimately seek the possibility of traversing Masking as much as possible. Through the data engine, we found that this is a very resource-saving method to achieve better training purposes. Combined with many of the most advanced methods, it greatly improves the sample quality during training.
\subsection{Salient Target Extraction}
Using the U²-Net-based prominent object extraction algorithm, we focused on container images. Initially, the SAM (Segment Anything Model) was employed for image collection and processing, resulting in a substantial dataset of container images along with their corresponding mask images for subsequent analysis. These images, along with their masks, were fed into the U²-Net for training, resulting in a prominent object detection model specifically designed for extracting containers from images.
\subsection{Container Morphology Compensation}
\begin{figure}[htbp]
    \centering
    \begin{minipage}{0.8\linewidth}
        \centering
        \includegraphics[width=0.9\linewidth]{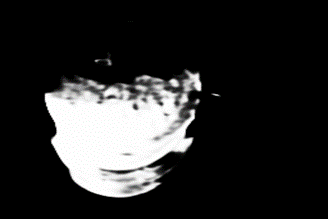}
        \caption{Before the Completion}
        \label{fig:before-the-completion}
    \end{minipage}

    \vspace{10pt}

    \begin{minipage}{0.8\linewidth}
        \centering
        \includegraphics[width=0.9\linewidth]{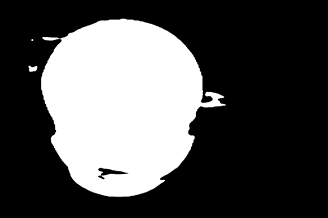}
        \caption{After the Completion}
        \label{fig:after-the-completion}
    \end{minipage}
\end{figure}

Following the application of the U²-Net model, certain images exhibited containers with colors closely resembling the surrounding environment, making them difficult to separate shown in Fig.~\ref{fig:before-the-completion}. This resulted in discontinuities between adjacent segmented images. To address this issue, morphological operations were applied to the images to fill in the gaps and obtain complete images, ensuring a stable and continuous segmentation of the images shown in Fig.~\ref{fig:after-the-completion}.
After being processed by the trained U²-Net salient target detection model, an image containing only the location in the image is obtained, and then fused with the original image to obtain an image containing only the container.
\subsection{Dynamic Liquid Level Detection}
In motion object detection using frame differencing, the goal is to detect the changing parts by eliminating the static regions and retaining the areas with variations in the difference image. Zhan et al. [6] divided the edge difference image into several small blocks and determined whether they were motion regions by comparing the number of non-zero pixels with a threshold. By applying this method, it is possible to extract the information about the changing liquid levels within the container.
\subsubsection{Threshold Division}
\ 

\begin{figure}[H]
    \centering
    \includegraphics[width=1\linewidth]{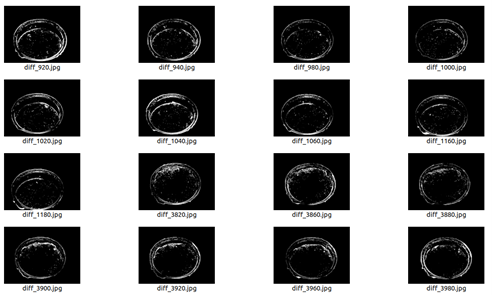}
    \caption{Threshold Division}
    \label{threshold division}
\end{figure}

\begin{figure}[H]
    \centering
    \includegraphics[width=1\linewidth]{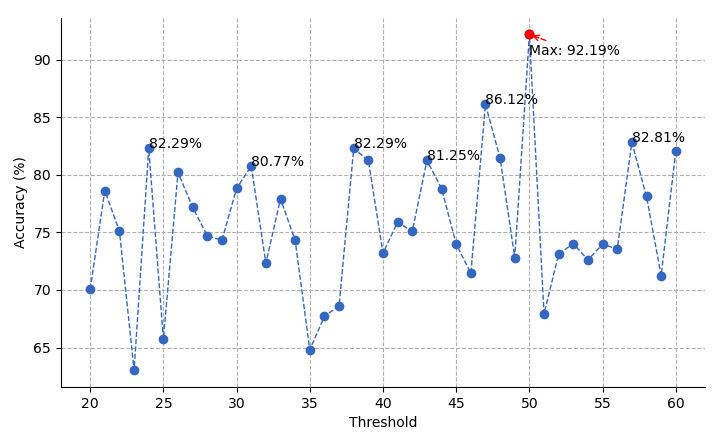}
    \caption{Threshold Data}
    \label{threshold digit}
\end{figure}

After converting the obtained container-only images to grayscale, the grayscale value difference between adjacent frames at corresponding pixel positions was calculated. A threshold value was established to partition the images according to these variations. Pixels with differences greater than the threshold were marked as white, while pixels with differences below the threshold were marked as black. This process captured subtle changes in the liquid level within the container shown in Fig.~\ref{threshold division} and assigned different labels to represent different liquid level states: no change in the lower level, rising level, no change in the higher level, falling level, and container movement. The labeled images were then fed into a neural network for image classification.
It is crucial to set a reasonable threshold that can clearly distinguish neighboring differences. Initially, we set the threshold range between 20 and 60 and experimented with the resulting difference images using different threshold values. The data in Fig.~\ref{threshold digit} shows the comparative results. The threshold value of 50 achieved the best performance, reaching \textbf{92.19\%}.
\subsubsection{Liquid level difference calculation}
\ 

The images of adjacent video frames are first converted from RGB to grayscale. While keeping the external environment and the container unchanged, variations in the grayscale values at corresponding positions between consecutive frames indicate subtle dynamic changes in the video sequence. We set a threshold for the magnitude of these value differences and determined the optimal threshold through experimental comparisons. Pixels at positions where the difference exceeds the threshold are marked. By processing these consecutive video frames, we obtain the dynamic changes in the liquid level within the container.
\subsubsection{Liquid level detection}
\ 

\begin{figure}[htbp]
	\centering
	\begin{minipage}{0.49\linewidth}
		\centering
		\includegraphics[width=0.9\linewidth]{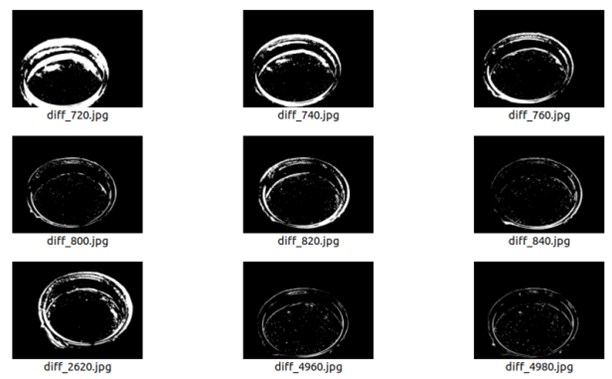}
		\caption{Increase}
		\label{increase}
	\end{minipage}
	\begin{minipage}{0.49\linewidth}
		\centering
		\includegraphics[width=0.9\linewidth]{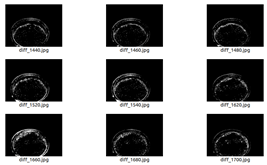}
		\caption{Decrease}
		\label{decrease}
	\end{minipage}
\end{figure}

Due to the training images being binary and the target object being relatively homogeneous, we selected a lightweight model such as EfficientNet-B0 for training. We fed these images shown in Fig.~\ref{increase} and Fig.~\ref{decrease} along with their corresponding labels into the neural network for image classification, resulting in a network capable of detecting images for this specific task. By utilizing this network, we ultimately achieved the detection of dynamic changes in liquid level.

\section{Experiment and result analysis}

Following our implementation of U²-Net for dynamic liquid level detection, we compared its performance with several well-known semantic segmentation models to benchmark its effectiveness. These models include U-Net\cite{ronneberger2015u}, DeepLabV3+\cite{chen2018encoder}, Mask R-CNN\cite{he2017mask}, F3Net\cite{wei2020f3net}, HRNet\cite{sun2019deep}, and PSPNet\cite{zhao2017pyramid}. The evaluation metrics employed in our comparison were Accuracy (Acc), Precision (P), Recall (R), F1-score, Mean Absolute Error (MAE), and Mean Squared Error (MSE).

\begin{table}[htbp]
\centering
\caption{Model Comparison}
\label{Model Comparison}
\begin{tabular}{@{}lcccccc@{}}
\toprule
Model        & Acc   & P     & R     & F1-score & MAE    & MSE    \\ 
\midrule
U-Net        & 0.977 & 0.694 & 0.784 & 0.736    & 10.529 & 49.121 \\
DeepLabV3+   & 0.979 & 0.757 & 0.817 & 0.767    & 10.956 & 29.384 \\
Mask R-CNN   & 0.968 & 0.763 & 0.836 & 0.788    & 0.852  & 0.781  \\
F3Net        & 0.974 & 0.748 & 0.823 & 0.771    & 9.874  & 10.321 \\
HRNet        & 0.981 & 0.774 & 0.809 & 0.785    & 0.874  & 0.974  \\
PSPNet       & 0.984 & 0.769 & 0.813 & 0.786    & 9.923  & 10.219 \\
\textbf{U²-Net}       & \textbf{0.991} & \textbf{0.794} & \textbf{0.848} & \textbf{0.812}    & \textbf{0.287}  & \textbf{0.002}  \\
\bottomrule
\end{tabular}
\end{table}

As indicated in Table \ref{Model Comparison}, U²-Net shows superior performance compared to the other models evaluated. It achieves the highest accuracy at 0.991, significantly higher than the next best-performing model, PSPNet, which has an accuracy of 0.984. U²-Net's precision and recall scores, 0.794 and 0.848 respectively, highlight its effectiveness in correctly classifying salient areas in the images.

The F1-score for U²-Net is 0.812, confirming its robustness and the effective balance it strikes between precision and recall. In terms of error metrics, U²-Net records the lowest values with a mean absolute error of 0.287 and a mean squared error of 0.002, emphasizing its precision and reliability in predicting liquid level changes.

These comparative results underscore the potential of U²-Net for practical deployment in scenarios where accurate liquid level detection is paramount, such as in industrial control systems. The evaluation suggests that U²-Net could serve as a reliable model for similar segmentation tasks that demand high accuracy and consistency.

\section{Conclusion}
In this study, we developed a novel approach for liquid level state detection by combining image differencing and binarization techniques. Our model demonstrated strong robustness against variations in container types and environmental conditions. By simplifying the input images into binary representations focusing on the target object, we were able to achieve accurate classification using a straightforward neural network architecture, without the need for complex network designs.

One of the key advantages of our model is its reduced reliance on large training datasets, which is a common challenge in many computer vision tasks. This was made possible by leveraging the SemiReward framework to generate and filter high-quality pseudo-labeled images using the SAM model. The resulting dedicated dataset enabled efficient training and generalization of our model.

The prospective uses of our methodology surpass the domain of liquid level state detection. The underlying principles can be adapted to a wide range of tasks that involve identifying small changes in static object environments. This versatility opens up opportunities for solving diverse problems across various domains, such as quality control in manufacturing, anomaly detection in surveillance systems, and monitoring of infrastructure conditions.

Integrating image differencing and object-focused binarization presents a potent approach for simplifying complex visual information into more manageable representations. By focusing on the essential features of the target object, our model can effectively capture and analyze the relevant changes while being resilient to variations of background. This approach not only enhances the robustness of the model but also reduces the computational complexity and data requirements, making it more practical for real-world deployments.

Furthermore, our model's ability to generate high-quality pseudo-labeled data using the SemiReward framework presents an opportunity for self-supervised learning. By iteratively refining the dataset and retraining the model, we can continuously improve its performance and adapt to new scenarios without the need for extensive manual labeling efforts. This self-supervised learning paradigm has the potential to greatly accelerate the development and deployment of computer vision models in various domains.

In conclusion, our liquid level state detection model, based on image differencing and binarization, offers a robust, efficient, and generalizable approach for analyzing small changes in static object environments. By simplifying complex images into binary representations and leveraging high-quality pseudo-labeled data, we have demonstrated the potential for solving a wide range of similar problems with reduced data requirements and computational complexity. As we continue to explore and refine this approach, we anticipate its application in diverse fields, contributing to advancements in computer vision and automation technologies.

\small
\bibliographystyle{unsrt}
\bibliography{main}
\normalsize

\end{document}